\documentclass[11pt,a4paper]{article}
\usepackage{authblk}  
\usepackage[hyperref]{emnlp2020}
\usepackage{times}
\usepackage{latexsym}

\usepackage[utf8x]{inputenc}
\usepackage{amsmath, amssymb}
\usepackage{mathrsfs}
\usepackage{graphicx}
\usepackage{booktabs}
\usepackage{soul}

\renewcommand{\to}{\rightarrow}

\newcommand{\code}{\texttt}
\let\cross\times

\renewcommand{\times}{\cdot}

\renewcommand{\bold}{\textbf}
\newcommand{\italic}{\textit}
\newcommand{\hidden}[2]{\hspace{-0em}#2}

\newcommand{\sub}[1]{_{\text{#1}}}
\newcommand{\loss}[1]{\ell\sub{#1}}

\renewcommand{\phi}{\varphi}
\renewcommand{\epsilon}{\varepsilon}

\renewcommand{\geq}{\geqslant}

\definecolor{brown}{RGB}{103, 04, 5}
\definecolor{lunar}{RGB}{21, 48, 122}
\definecolor{Gray}{gray}{0.9}
\definecolor{mint}{rgb}{0.24, 0.71, 0.54}

\usepackage{amsthm}
\usepackage{enumerate}
\theoremstyle{definition}

\usepackage[normalem]{ulem} 
\usepackage[g]{esvect} 

\usepackage{multirow}  
\usepackage{color, colortbl}
\usepackage{subcaption}

\usepackage{microtype}

\aclfinalcopy 
\def\aclpaperid{2920} 

\title{Are ``{U}ndocumented {W}orkers" the {S}ame as ``{I}llegal {A}liens"? {D}isentangling {D}enotation and {C}onnotation in {V}ector {S}paces} 

\author[1,2]{Albert Webson}
\author[3]{Zhizhong Chen}
\author[1]{Carsten Eickhoff}
\author[1]{Ellie Pavlick}
\affil[ ]{\{albert\_webson, zhizhong\_chen, carsten, ellie\_pavlick\}@brown.edu}
\affil[1]{Department of Computer Science, Brown University}
\affil[2]{Department of Philosophy, Brown University}
\affil[3]{Department of Physics, Brown University}

\begin{document}

\maketitle

\begin{abstract}
In politics, neologisms are frequently invented for partisan objectives. For example, “undocumented workers” and “illegal aliens” refer to the same group of people (i.e., they have the same denotation), but they carry clearly different connotations. Examples like these have traditionally posed a challenge to reference-based semantic theories and led to increasing acceptance of alternative theories (e.g., Two-Factor Semantics) among philosophers and cognitive scientists. In NLP, however, popular pretrained models encode both denotation and connotation as one entangled representation. In this study, we propose an adversarial neural network that decomposes a pretrained representation as independent denotation and connotation representations. For intrinsic interpretability, we show that words with the same denotation but different connotations (e.g.,  “immigrants” vs.\ “aliens”, “estate tax” vs.\ “death tax”) move closer to each other in denotation space while moving further apart in connotation space. For extrinsic application, we train an information retrieval system with our disentangled representations and show that the denotation vectors improve the viewpoint diversity of document rankings. 
\end{abstract}

\section{Introduction}
\label{sec:intro}

\begin{figure}[ht!]
\includegraphics[width=\linewidth]{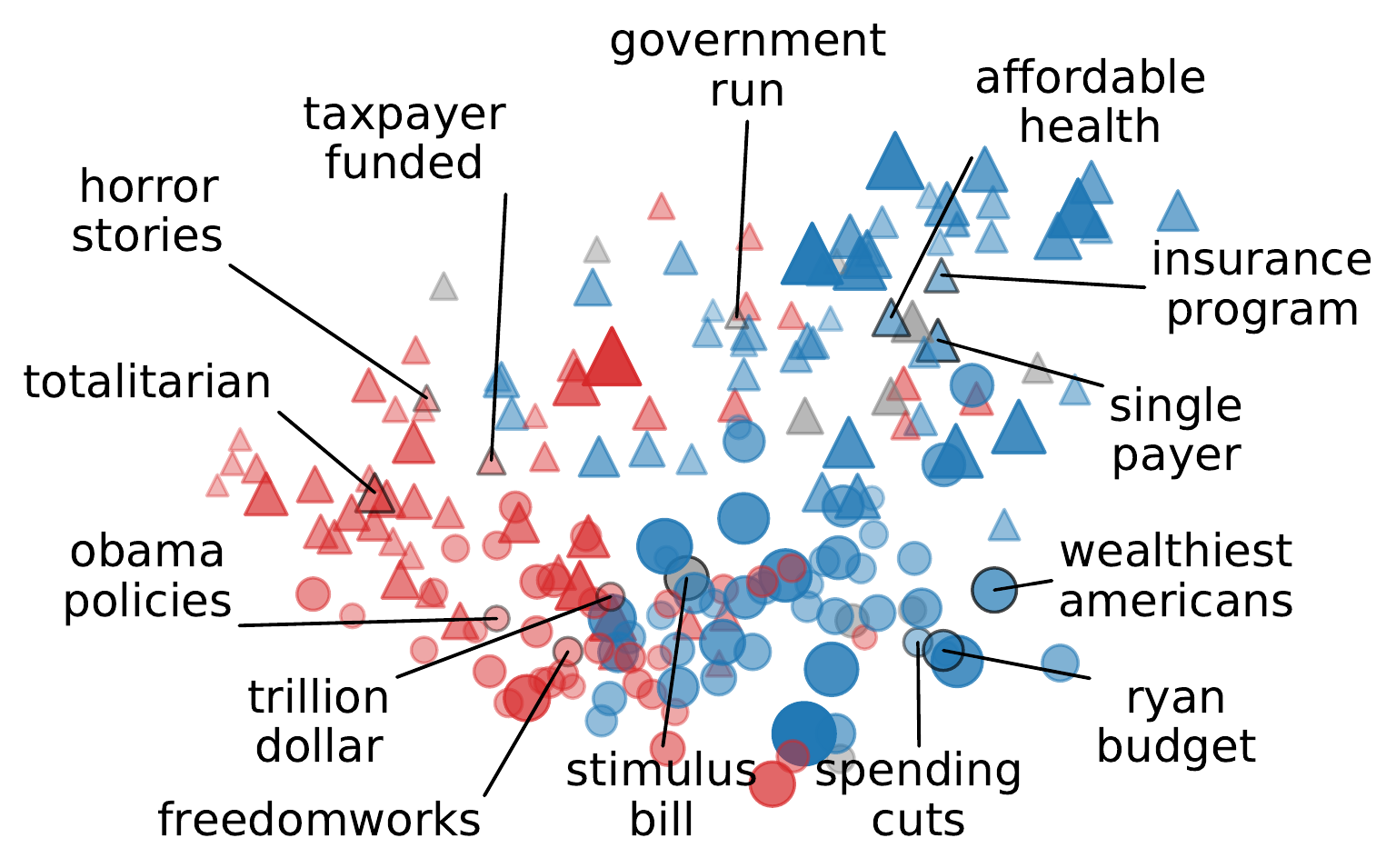}
\caption{Nearest neighbors of government-run healthcare (triangles) and economic stimulus (circles). Note that words cluster as strongly by policy denotation (shapes) as by partisan connotation (colors); namely, pretrained representations conflate denotation with connotation. Plotted by t-SNE with perplexity = 10.}
\label{fig:motivating}
\end{figure}

Language carries information through both denotation and connotation. For example, a reporter writing an article about the leftmost wing of the Democratic party can choose to refer to the group as ``progressives'' or as ``radicals''. The word choice does not change the individuals referred to, but it does communicate significantly different sentiments about the policy positions discussed. This type of linguistic nuance presents a significant challenge for natural language processing systems, most of which fundamentally assume words to have similar meanings if they are surrounded in similar word contexts. Such assumption risks confusing differences in connotation for differences in denotation or vice versa. For example, using a common skip-gram model \citep{Mikolov2013} trained on a news corpus (described in $\S\ref{sec:news-corpus}$), Figure~\ref{fig:motivating} shows nearest neighbors of ``government-run healthcare'' and ``economic stimulus''. The resulting t-SNE clusters are influenced as much by policy denotation (shapes) as they are by partisan connotation (colors\footnote{Throughout this paper, blue reflects partisan leaning toward the Democratic Party and red reflects partisan leaning toward the Republican Party in the United States.}). Using these entangled representations in applications such as information retrieval could have pernicious consequences such as reinforcing ideological echo chambers and political polarization. For example, a right-leaning query like ``taxpayer-funded healthcare" could make one equally (if not more) likely to see articles about ``totalitarian'' and ``horror stories'' than about ``affordable healthcare''. 

To address this, we propose classifier probes that measure denotation and connotation information in a given pretrained representation, and we arrange the probe losses in an adversarial setup in order to decompose the entangled pretrained meaning into distinct denotation and connotation representations (\S\ref{sec:model}). We evaluate our model intrinsically and show that the decomposed representations effectively disentangle these two dimensions of semantics (\S\ref{sec:intrinsic}). We then apply the decomposed vectors to an information retrieval task and demonstrate that our method improves the viewpoint diversity of the retrieved documents (\S\ref{sec:extrinsic}). All data, code, preprocessing procedures, and hyperparameters are included in the appendix and our GitHub repository.\footnote{https://github.com/awebson/congressional\_adversary}

\section{Philosophical Motivation}
\label{sec:formulation} 

Consider the following two sentences: ``Undocumented workers are undocumented workers" vs. ``Undocumented workers are illegal aliens''. \citet{frege1892sinn} famously used sentence pairs like these, which have the same truth conditions but clearly different meanings, in order to argue that meaning is composed of two components: ``reference'', which is some set of entities or state of affairs, and ``sense'', which accounts for how the reference is presented, encompassing a large range of aspects such as speaker belief and social convention.

In contemporary philosophy of language, the sense and reference argument has evolved into debates of semantic externalism vs.\ internalism and referential vs.\ conceptual role semantics. Externalists and referentialists\footnotemark\ continue the truth-conditional tradition and emphasize meaning as some entity to which one is causally linked, invariant of one's psychological encoding of the referent \citep{putnam1975meaning, kripke1972naming}. On the other hand, conceptual role semanticists emphasize meaning as what inferences one can draw from a lexical concept, deemphasizing the exact entities which the concept includes \citep{greenberg2005conceptual}. Naturally, a popular position takes the Cartesian product of both schools of meaning \citep{Block1986, Carey2009}. This view is known as Two-Factor Semantics, and it forms the inspiration for our work. To avoid confusion with definitions from existing literature, we use the terms ``denotation'' and ``connotation'' rather than ``reference'' and ``concept'' when discussing our models in this paper.  \footnotetext{Technically, one can be a referentialist while also being a semantic internalist. See \citet{sep-word-meaning} for a nuanced overview as well as related theories in linguistics and cognitive science.}

\section{Data}  
\label{sec:data}

 We assume that it is possible to disentangle the two factors of semantics by grounding language to different components of the non-linguistic context. In particular, our approach assumes access to a set of training sentences, each of which grounds to a denotation $d$ (which approximates reference) or a connotation $c$ (which approximates conceptual inferences). We require at least one of $d$ or $c$ to be observed, but we do not require both (elaborated in $\S$\ref{sec:deno-labels}). In this work, $d$ and $c$ are discrete symbols. However, our model could be extended to settings in which $d$ and $c$ are feature vectors. 

While we are interested in learning lexical-level denotation and connotation, we train on sentence- and document-level speaker and reference labels. We argue that this emulates a more realistic form of supervision. For example, we often have metadata about a politician (e.g., party and home state) when reading or listening to what they say, and we are able to aggregate this to make lexical-level judgements about denotation and connotation. 

\begin{figure}[ht!]
    \centering
    \includegraphics[width=\linewidth]{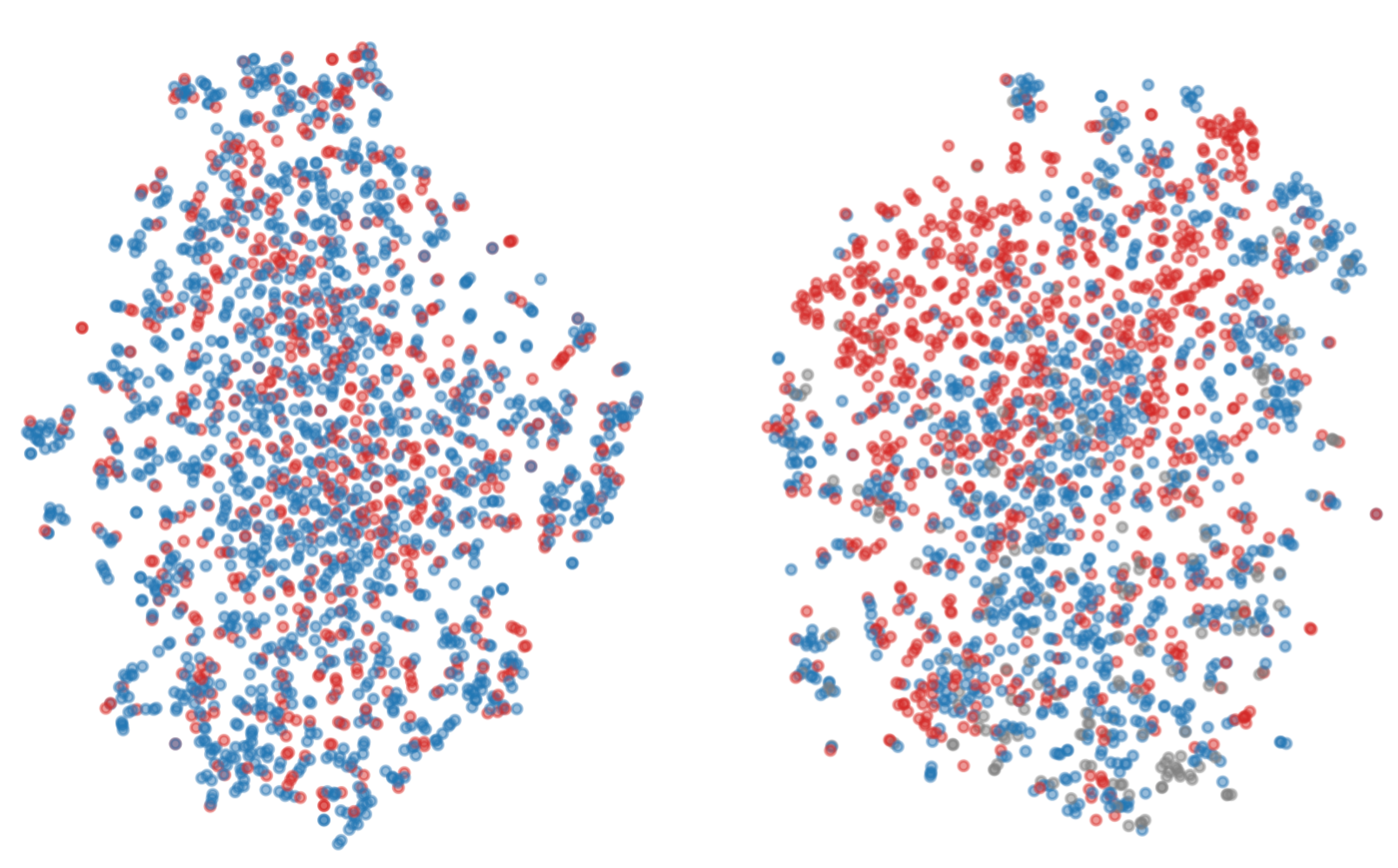}
    \caption{Vector spaces that result from training vanilla word2vec on the Congressional Record (left) and Partisan News (right). We evaluate on both corpora, but note that the Partisan News corpus better exemplifies the problem we target where words cluster strongly according to ideological stance.}
    \label{fig:large-clusters}
\end{figure}

We experiment on two corpora: the Congressional Record (CR) and the Partisan News Corpus (PN), which differ in linguistic style, partisanship distribution (Figure~\ref{fig:large-clusters}), and the available labels for grounding denotation and connotation. 

\begin{table*}[ht!]
\centering
\resizebox{\linewidth}{!}{
\begin{tabular}{llrrll}
\toprule
Name  & Corpus & Vocab. & Num. Sent. & Denotation Grounding & Connotation Grounding \\
\midrule
\textsc{CR Bill} & Congr. Record & 21,170 & 381,847 & legislation title (1,029-class) & speaker party (2-class)\\
 \textsc{CR Topic} & Congr. Record & 21,170 & 381,847 & policy topic (41-class) & speaker party (2-class)\\
 \textsc{CR Proxy}  & Congr. Record & 111,215 & 5,686,864 & none (LM proxy) & speaker party (2-class)\\
\textsc{PN Proxy}   & Partisan News & 138,439 & 3,209,933 & none (LM proxy) & publisher partisan leaning (3-class)\\
\bottomrule
\end{tabular}}
\caption{Summary of model variants experimented.}
\label{tab:model_variants}
\end{table*}

\subsection{Congressional Record}
\label{sec:cr}

The Congressional Record (CR) is the official transcript of the floor speeches and debates of the United States Congress dating back to 1873. \citet{gentzkow2019measuring} digitized and identified approximately 70\% of these speeches with a unique speaker, where each speaker is labeled with their gender, party, chamber, state, and district. To constrain the political and linguistic change over time, we use a subset of the corpus from 1981 to 2011.\footnote{2011 is the latest session available for the Bound Edition of CR; 1981 is chosen because the Reagan Administration marks the last party realignment and thus we can expect connotation signals to remain reasonably consistent over this period.}

In order to assign labels that can be used as proxies of denotation, we weakly label each sentence with both its legislative topic and the specific bill being debated.\footnote{\label{hard-label}We also experimented with collecting more precise reference labels using the entity linkers of both Google Cloud and Facebook Research on a variety of corpora. However, the results of entity linking were too poor to justify pursuing this direction further. We would love to see future works that devise creative ways to include better denotation grounding.} To do this, we collected a list of congressional bills from the U.S. Government Publishing Office.\footnote{https://www.govinfo.gov/bulkdata/BILLSTATUS} For our purposes, this data provides the congressional session, policy topic, and an informal short title for each bill. We perform a regular expression search for each bill’s short title among the speeches in its corresponding congressional session. For bills that are mentioned at least 3 times, we assume that the speech in which the bill was mentioned as well as 3 subsequent speeches are referring to that bill, and we label each speech with the title and the policy topic of that bill. Speeches that are not labeled by this process are discarded. Additional details and examples are given in Appendix \ref{apd:bill-mentions}.

\subsection{Partisan News Corpus}
\label{sec:news-corpus}

Hyperpartisan News is a set of web articles collected for a 2019 SemEval Task  \citep{kiesel-etal-2019-semeval}. It consists of articles scraped from the political sections of 383 news outlets in English. Each article is associated with a publisher which, in turn, has been manually labeled with a partisan leaning on a five-point scale: ``left, center-left, center, center-right, right". Upon manual inspection, we find that the distinctions between right vs. center-right and left vs. center-left are prone to annotation artifacts. Therefore, we collapse these labels into a three-point scale, and we refer to this 3-class corpus as the Partisan News (PN) corpus throughout. No denotation label is available for this corpus.

\begin{figure*}[ht!]
    \centering
    \includegraphics[width=15cm]{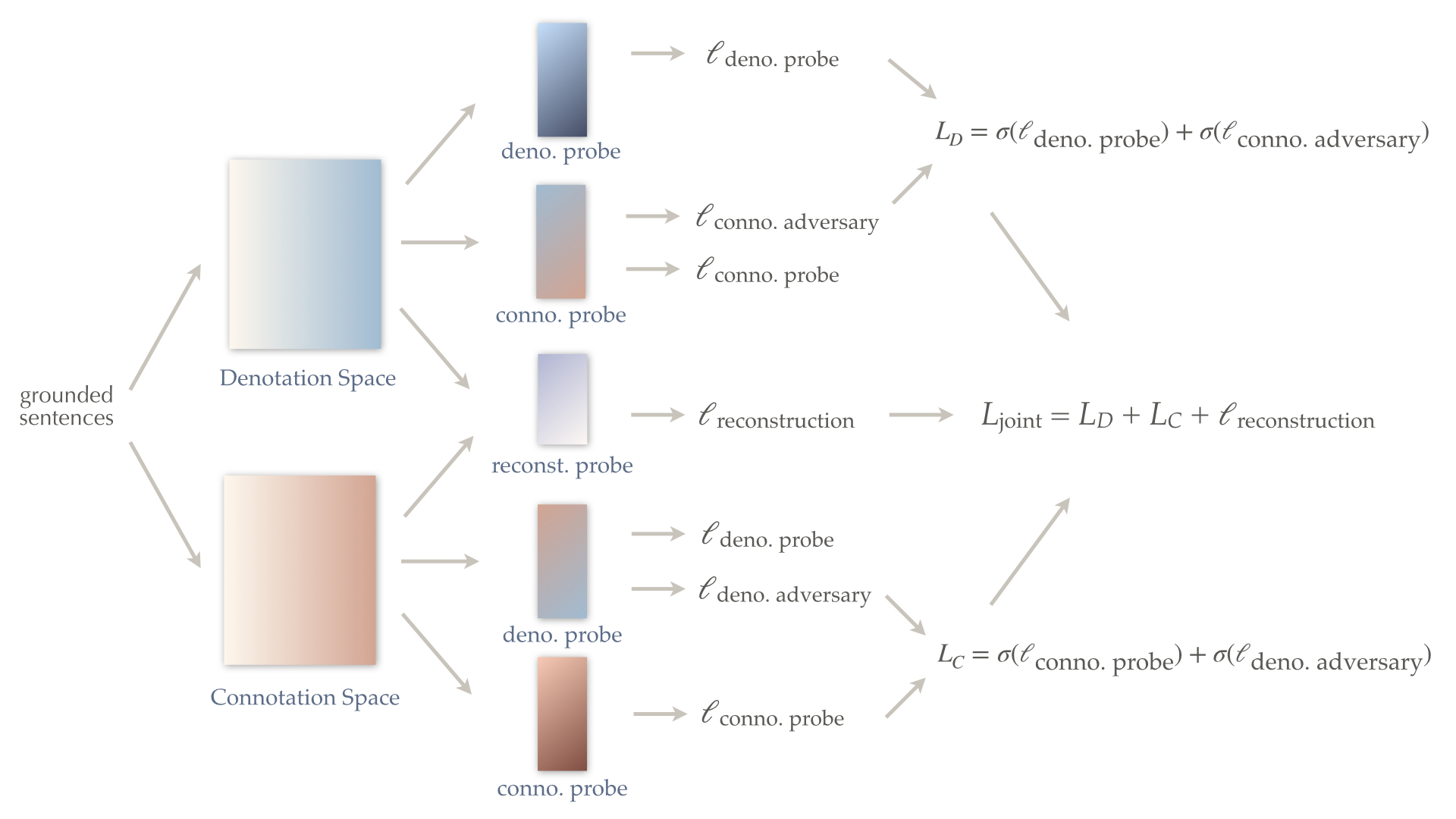}
    \caption{Overall model and composition of losses}
    \label{fig:network}
\end{figure*}

\section{Model}
\label{sec:model}

Section \ref{sec:general-arch} describes our model architecture. Sections~\ref{sec:cono-labels} and \ref{sec:deno-labels} then describe specific instantiations that we use in our experiments. These variants are summarized in Table~\ref{tab:model_variants}.

\subsection{Overall Architecture}
\label{sec:general-arch}

Let $V\sub{deno}, V\sub{conno}, V\sub{pretrained}$ be the vector spaces of denotation, connotation, and pretrained spaces respectively. Our model consists of two adversarial decomposers: 
\begin{align*}
    D&: V\sub{pretrained} \to V\sub{deno} \\
    C&: V\sub{pretrained} \to V\sub{conno}
\end{align*}
The goal is to train $D$ to \italic{preserve} as much denotation information as possible while \italic{removing} as much connotation information as possible from the pretrained representation. Symmetrically, $C$ will preserve as much \italic{connotation} as possible while removing as much \italic{denotation} as possible from the pretrained representation.

For clarity, let us focus on $D$ for now. To measure how much denotation or connotation structure is encoded in $V\sub{deno}$, we use two classifiers probes trained to predict the denotation label $d$ or connotation label $c$, which yield two cross-entropy losses $\loss{deno. probe}$ and $\loss{cono. probe}$ respectively. In order to encourage the decomposer $D$ to preserve denotation and remove connotation, we define its loss function as $$L_D = \sigma(\loss{deno. probe}) + \sigma(\loss{conno. adversary})$$
where $\sigma$ is the sigmoid function and
\begin{align*}
\loss{conno. adversary} = \text{KL Div}\, ( &\text{conno. probe predicted dist.}, \\&\text{uniform dist.} )
\end{align*}
The adversarial loss $\loss{conno. adversary}$ rewards $D$ to remove connotation structure such that the probe prediction is random. Meanwhile, the probes themselves are still only gradient updated with the usual cross-entropy losses—extracting and measuring as much denotation or connotation information as possible—independent of the decomposer $D$. 

As shown in Figure \ref{fig:network}, $C$ is set up symmetrically, so it is trained with the usual classification loss from its connotation probe and a KL divergence adversarial loss from its denotation probe.

Finally, we impose a reconstruction probe $R$ with the loss function:
\begin{align*}
    \loss{recon.} = 1 &- \text{cos sim} (R(v\sub{deno}, v\sub{conno}), v\sub{pretrained}) 
\end{align*}
which enforces that the combination of denotation and connotation subspaces preserves all the semantic meaning of the original pretrained space, as opposed to merely encoding predictive features that maximize probe accuracies. (We verified in ablation experiments that this is in fact what happens without $R$.) Assembling everything together, the decomposers $D$ and $C$ are jointly trained with $L\sub{Joint} = L_{D} + L_{C} +  \loss{recon.}$.

In principle, $D$ and $C$ can be a variety of sentence encoders. In this work, we implement them as simple mean bags of static embedding for two reasons: First, it is difficult to interpret contextualized embedding for an individual word (especially for the type of analysis we present in \S\ref{sec:intrinsic}). Second, many of the interesting heavily connotative expressions consist of multiple words (e.g., ``socialized medicine", ``universal healthcare'') and compositionality is still far from being solved. Therefore, we conjoin multiword expressions with underscores so that we can model them in the same way as atomic words.\footnote{Appendix \ref{apd:preproc} documents this preprocessing step in detail. Throughout this paper, ``words'' refers to both individual words and underscored short phrases.}

\subsection{Connotation Probes}
\label{sec:cono-labels}

We exploit the fact that much of the debate in American politics today is (sadly) reducible to partisan division \citep{lee2009beyond, klein2020we}, thus it is safe to define the connotation label of every document to be simply the partisanship of the speaker. Of course, connotation in the general domain can encompass much more than liberals vs.\ conservatives, and in future work, we hope to extend this to multi-faceted connotations that are more true to the semantic theories as discussed in \S\ref{sec:formulation}. For now, in CR, connotation is the speaker's party, and in PN, connotation is the partisan leaning of the publisher. 

 Again, in principle, the probes can be a variety of neural modules. In this work, we implement the connotation probes as 4-layer MLPs. We experimented with the more popular 1-layer MLP and 1-linear-layer probes. However, when the probes are shallow, the model converges before most of the information that should be removed is in fact removed. For example, when we use a 4-layer MLP probe on a decomposed representation trained with a 1-layer probe, the 4-layer probe accuracies are just as good as if the representation has not been decomposed at all. That is, our experiments suggest that the probes have to be sufficiently complex in order to truly measure what denotation/connotation structure is removed or preserved in a decomposed representation. 

\subsection{Denotation Probes}
\label{sec:deno-labels}

For the CR corpus, we experimented with two types of denotation labels: The specific piece of legislation under discussion and the general policy topic under discussion. In \textsc{CR Bill}, the label is one of the 1,029 short titles of bills. In \textsc{CR Topic}, the label is one of 41 policy topics. Both types of labels are annotated as described in $\S$\ref{sec:cr}. For the same reason as discussed in the previous paragraph, we implement the denotation probes as 4-layer MLPs.

Additionally, as mentioned in Footnote \ref{hard-label}, precise denotation labels are difficult to collect, so we also experimented with more realistic settings (\textsc{CR Proxy} and \textsc{PN Proxy}) which do not use any denotation labels. In this case, we return to the theories discussed in \S\ref{sec:formulation} and note that, because semantic meaning can be partitioned into two components, we may assume pretrained representations encode the overall meaning and any aspects of meaning that are not explained by our connotation labels must belong to denotation.\footnotemark\ Thus, we may continue to use the pretraining objective (in this implementation, skip-gram-style context word prediction) as a proxy probe for denotation information and rely on the adversarial connotation probe to remove connotation structure in the denotation space. 

\footnotetext{We acknowledge that this feels a bit backward: Ideally, in a Fregean sense, everything not explained by reference is left over to sense, rather than the converse. However, we are constrained by the available grounding. In a different setting, if we had explicit referential labels but no speaker information, we could use skip-gram as the proxy for connotation instead.} 

\begin{figure*}[ht!]
    \centering
        \includegraphics[width=\linewidth]{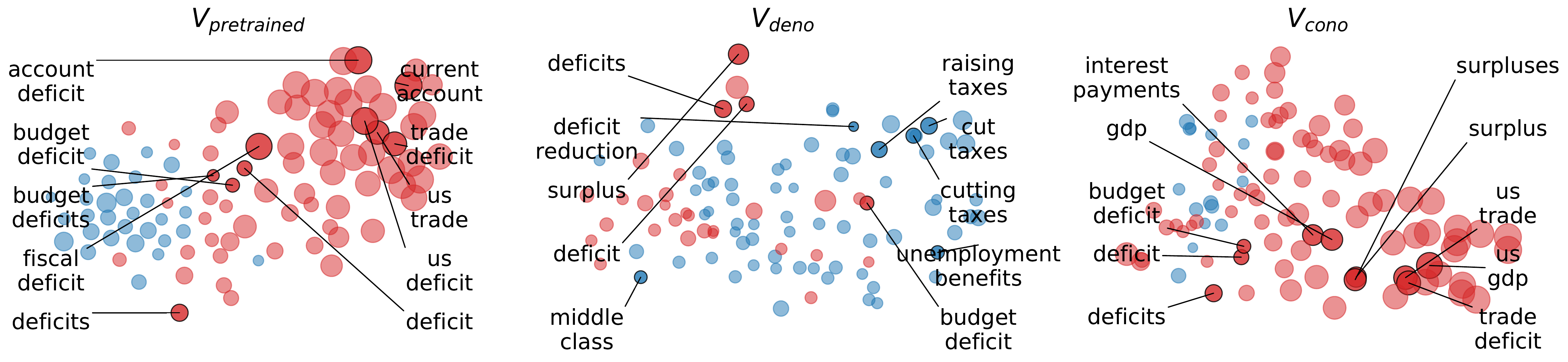}
    \caption{Neighborhood of ``deficit'' in $V\sub{pretained}$, $V\sub{deno}$, and $V\sub{conno}$ of $\textsc{PN Proxy} $. Arrows point to the top-10 nearest neighbors. Colors reflect partisan leaning, where more opaque dots are more heavily partisan words. Note that in $V\sub{pretained}$ and in $V\sub{conno}$, the nearest neighbors are all Republican-leaning words, whereas they are balanced in $V\sub{deno}$.}
    \label{fig:clusters}
\end{figure*}

\section{Intrinsic Evaluation}
\label{sec:intrinsic}
We confirm that our decomposed denotation and connotation spaces reflect their intended purposes by measuring their structures with homogeneity metrics ($\S$\ref{sec:homo-metric}) on three sets of evaluation words ($\S$\ref{sec:test-sets}) as well as inspecting their t-SNE clusters. 

\subsection{Homogeneity Metrics}
\label{sec:homo-metric}
To quantify how much denotation or connotation structure is encoded in a vector space, we define the homogeneity ($h\sub{deno}$, $h\sub{conno}$) of a given space to be the average proportion of a query word's top-$k$ nearest neighbors\footnote{We set $k=10$, but we found that evaluation results remain robust across different choices of $k$.} which share the same denotation/connotation label as the query's own denotation/connotation label.\footnote{We also ran \texttt{sklearn.homogeneity\_score} but saw no difference in trends, so we report our homogeneity metric for its simple interpretability.} In particular, we are interested in comparing the delta of $V\sub{deno}$ and $V\sub{conno}$ against $V\sub{pretrained}$. For $V\sub{deno}$, we hope to see $h\sub{deno}$ \textit{increase} relative to the pretrained space and see $h\sub{conno}$ \textit{decrease} relative to the pretrained space. For $V\sub{conno}$, we hope to observe movement in the opposite direction.

As motivated in $\S$\ref{sec:data}, our model is trained with labels at the sentence-level, while homogeneities are evaluated at the word-level. We assign a word's connotation label to simply be the party that uses the word most often. For \textsc{CR Bill} and \textsc{CR Topic}, we assign the word-level denotation label as either the bill or the topic that uses the word most often. For the PN corpus, no ground truth denotation label is available, so we cannot directly measure $h\sub{deno}$, but we show alternative evaluation in $\S$\ref{sec:intrinsic-results}. Table \ref{tab:baselines} shows the baseline $h\sub{deno}$ and $h\sub{conno}$ scores for embeddings pretrained on each corpus and evaluating over two test sets of words (described in the next section).

\begin{table*}[ht!]
\centering
\small
\begin{tabular}{llcccccccc}
\toprule
	&& 				\multicolumn{4}{c}{$V\sub{deno}$ (and $\Delta$ with $V\sub{pre}$) }	&	\multicolumn{4}{c}{$V\sub{conno}$  (and $\Delta$ with $V\sub{pre}$)} \\		
\cmidrule(lr){3-6} \cmidrule(lr){7-10}
Test Set &	 Model & $h\sub{deno}$ & $\Delta$ ($\uparrow$) & $h\sub{conno}$ & $\Delta$ ($\downarrow$) & $h\sub{deno}$ & $\Delta$ ($\downarrow$) & $h\sub{conno}$ & $\Delta$ ($\uparrow$)\\
\cmidrule(lr){1-1}	\cmidrule(lr){2-2} \cmidrule(lr){3-4} 	\cmidrule(lr){5-6} 	 	\cmidrule(lr){7-8} 	\cmidrule(lr){9-10} 	
High Partisan &

\textsc{CR Bill}
&	$0.28$		&	$+0.09$		
&	$0.65$		&	$-0.11$		
&	$0.02$		&	$-0.17$		
&	$0.89$		&	$+0.13$	\\

&\textsc{CR Topic} 	
&	$0.53$		&	$+0.18$		
&	$0.59$		&	$-0.17$		
&	$0.07$		&	$-0.28$		
&	$0.98$		&	$+0.21$	\\

& \textsc{CR Proxy} 		
&	$0.07$		&	$+0.00$		
&	$0.71$		&	$-0.00$		
&	$0.04$		&	$-0.03$		
&	$0.99$		&	$+0.28$		\\

& \textsc{PN Proxy} 		
&	--		&	--		
&	$0.40$		&	$-0.26$				
&	--		&	--	
&	$0.76$		&	$+0.10$		\\

\midrule
Random 
& \textsc{CR Bill}
&	$0.14$		&	$+0.05$		
&	$0.69$		&	$-0.01$		
&	$0.04$		&	$-0.06$		
&	$0.77$		&	$+0.07$	\\

& \textsc{CR Topic}
&	$0.31$		&	$+0.02$		
&	$0.63$		&	$-0.07$		
&	$0.14$		&	$-0.15$		
&	$0.81$		&	$+0.11$	\\

& \textsc{CR Proxy} 		
&	$0.04$		&	$+0.00$		
& 	$0.64$		&	$-0.00$		
&	$0.02$		&	$-0.03$			
&	$0.85$		&	$+0.21$		\\

& \textsc{PN Proxy} 		
&	--		&	--		
&	$0.39$		&	$-0.21$				
&	--		&	--	
&	$0.69$		&	$+0.09$		\\
\bottomrule
\end{tabular}
\caption{Intrinsic evaluation results across models and test sets. $\Delta$ is change relative to $V\sub{pretrained}$ (Table \ref{tab:baselines}). Arrows in parentheses mark the desired directions of change. Note that because denotation labels have far more classes than connotation labels, the magnitude of $h\sub{deno}$ and $h\sub{conno}$ are not directly comparable with each other.}
\label{tab:intrinsic}
\end{table*}

\begin{table}[ht!]
\centering
\small
\begin{tabular}{lllll}
\toprule
			& \multicolumn{2}{c}{High Partisan}		& \multicolumn{2}{c}{Random} \\
			& $h\sub{deno}$		&	$h\sub{conno}$	& $h\sub{deno}$		&	$h\sub{conno}$\\
\cmidrule(lr){2-3} \cmidrule(lr){4-5}
\textsc{CR Bills}		&	0.19	&	0.76	&		0.09	&	0.70	\\
\textsc{CR Topic}		&	0.35	&	0.76	&	    0.29	&	0.70	\\
\textsc{CR Proxy} 		&	0.07	&	0.71	&	0.05	&	0.64	\\
\textsc{PN Proxy} 		&	--		&	0.66	&		--		&	0.60	\\
\bottomrule
\end{tabular}
\caption{Baseline homogeneity scores of embeddings pretrained on each corpus.}
\label{tab:baselines}
\end{table}

\subsection{Test Sets}
\label{sec:test-sets}

We evaluate on words sampled in three different ways: \textbf{Random} is a random sample of 500 words drawn from each corpus' vocabulary that occur at least 100 times in order to filter out web scraping artifacts, e.g., URLs and author bylines. 
\textbf{High Partisan} is a sample of around 300 words from each corpus's vocabulary that occur at least 100 times and have high partisan skew; namely, words that are uttered by a single party more than 70\% of the time. This threshold is chosen based on manual inspection, but we have evaluated on other thresholds as well with no significant difference in results. This High Partisan set is then bisected into two disjoint sets as dev and test data for model selection. All word sets sampled at different ratios are included in our released data. Finally, \bold{Luntz-esque} is a small set of manually-vetted pairs of words that are known to have the same denotation but different connotations. Most of them are drawn from \italic{The New American Lexicon} (Luntz 2006\footnote{This is a leaked report circulated via a Google Drive link which has been taken offline since. A copy is included in our released data.}), a famous report from focus group research which explicitly prescribes word choices that are empirically favorable to the Republican party line.

\begin{table*}[ht!]
\centering
\resizebox{\textwidth}{!}{ 
\begin{tabular}{l|rrr|rrr|rrr|rrr}
\toprule
			& \multicolumn{3}{c|}{\textsc{CR Bill}}
			& \multicolumn{3}{c|}{\textsc{CR Topic}}
			&\multicolumn{3}{c|}{\textsc{CR Proxy} } 
			& \multicolumn{3}{c}{\textsc{PN Proxy} }	\\
			& $V_{pre}$ & $\Delta V_d (\uparrow)$ & $\Delta V_c (\downarrow)$
			& $V_{pre}$ & $\Delta V_d (\uparrow)$ & $\Delta V_c (\downarrow)$
			& $V_{pre}$ & $\Delta V_d (\uparrow)$ & $\Delta V_c (\downarrow)$
			& $V_{pre}$ & $\Delta V_d (\uparrow)$ & $\Delta V_c (\downarrow)$ \\
\cmidrule(lr){1-1}\cmidrule(lr){2-4} \cmidrule(lr){5-7} \cmidrule(lr){8-10} \cmidrule(lr){11-13}
undocumented workers/illegal aliens & $0.81$ & $+0.03$ & $-0.01$ & $0.81$ & $-0.09$ & $+0.14$ &$0.95$ & $+0.03$ & $-1.28$ & $0.96$ & $+0.01$ & $-0.20$\\
estate tax/death tax & $0.89$ & $+0.05$ & $-0.76$ & $0.89$ & $+0.08$ & $-0.84$ & $0.96$ & $+0.00$ & $-0.98$ & $0.93$ & $+0.01$ & $-0.06$\\
capitalism/free market & $0.79$ & $+0.11$ & $+0.03$ & $0.79$ & $+0.14$ & $+0.16$ & $0.85$ & $-0.07$ & $-0.20$ & $0.96$ & $-0.01$ & $-0.02$\\
foreign trade/international trade & $0.90$ & $-0.05$ & $+0.02$ & $0.90$ & $+0.02$ & $-0.01$ & $0.86$ & $+0.05$ & $-0.40$ & $0.93$ & $+0.03$ & $+0.00$\\
public option/government-run & $0.67$ & $+0.06$ & $-0.57$ & $0.67$ & $+0.24$ & $-0.84$ & $0.92$ & $+0.02$ & $-1.08$ &$0.97$ & $+0.00$ & $-0.01$\\
trickle-down/cut taxes & -- & -- &-- & -- & -- & -- & $0.87$ & $+0.02$ & $-0.51$ & $0.95$ & $+0.02$ & $-0.12$\\
voodoo economics/supply-side & -- & -- &-- & -- & -- &-- & $0.95$ & $-0.04$ & $-0.07$ & $0.91$ & $+0.05$ & $-0.05$\\
tax expenditures/spending programs & -- & -- &-- & -- & -- &-- & $0.93$ & $-0.17$ & $-1.03$ & $0.99$ & $+0.00$ & $-0.16$\\
waterboarding/interrogation & -- & -- &-- & -- & -- &-- & $0.90$ & $-0.04$ & $-0.22$ & $0.97$ & $+0.01$ & $-0.01$\\
socialized medicine/single-payer & -- & -- &-- & -- & -- &-- & $0.88$ & $-0.11$ & $-0.56$ & $0.89$ & $+0.02$ & $-0.03$\\
political speech/campaign spending & -- & -- &-- & -- & -- & -- & $0.86$ & $-0.02$ & $-0.81$ & $0.99$ & $+0.00$ & $-0.05$\\
star wars/strategic defense initiative & -- & -- &-- & -- & -- &-- & $0.91$ & $-0.16$ & $-0.69$ & -- & -- &--\\
nuclear option/constitutional option & -- & -- &-- & -- & -- &-- & $0.97$ & $-0.14$ & $-1.30$ & -- & -- &--\\
\midrule
Changes in the Correct Direction & & $4/5$ & $3/5$ & & $4/5$ & $3/5$  & & $5/13$ & $13/13$  & & $10/11$  & $10/11$ \\
\bottomrule
\end{tabular}}
\caption{Changes in cosine similarity (relative to $V\sub{pretrained}$) for known political euphemism' pairs, i.e. words with the same denotation but opposite partisan connotation. Omitted entries are out of vocabulary.} 
\label{tab:luntz}
\end{table*}

\subsection{Results} 
\label{sec:intrinsic-results}

Overall, we see that our $V\sub{deno}$ and $V\sub{conno}$ spaces demonstrate the desired shift in homogeneities and structures, which is intuitively illustrated by Figure~\ref{fig:clusters}. Quantitatively, Table~\ref{tab:intrinsic} enumerates the homogeneity scores of both decomposed spaces as well as their directions of change relative to the pretrained space. For $V\sub{deno}$, we see that denotation homogeneity $h\sub{deno}$ consistently increases and connotation homogeneity $h\sub{conno}$ consistently decreases as desired. Conversely, for $V\sub{conno}$, we see $h\sub{conno}$ increases and $h\sub{deno}$ decreases as desired. Further, we see that the magnitude of change is greater across the board for the highly partisan words than for random words, which is expected as the highly partisan words are usually loaded with more denotation or connotation information that can be manipulated. The only exception is \textsc{CR Proxy}'s $V\sub{deno}$, which sees no significant movement in either direction. This is understandable because \textsc{CR Proxy} is not trained with ground truth denotation labels. (We evaluate it with the labels from \textsc{CR Bill}). 

As means of closer inspection, we compute the cosine similarities of words in our Luntz-esque analysis set. Because these pairs of words are known to be political euphemisms (e.g. “estate tax” and “death tax”, which refer to the same tax policy but imply opposite partisanship), we expect these pairs to become more cosine similar in $V\sub{deno}$ and less cosine similar in $V\sub{conno}$. As shown in Table~\ref{tab:luntz}, even without ground truth denotation labels, the $V\sub{deno}$ of \textsc{CR Proxy} and \textsc{PN Proxy} still preserve the pretrained denotation structure reasonably well. For pairs that do see decrease in $V\sub{deno}$  similarity, the errors are far smaller relative to their correct reduction in $V\sub{conno}$ similarity. For example, “political speech” and “campaign spending” experience a small ($-0.02$) decrease in denotation similarity; in exchange, the model correctly recognizes that the two words have opposite ideologies ($-0.81$ in connotation similarity) on the issue of whether unlimited campaign donation is shielded by the First Amendment as ``political speech". 

\section{Extrinsic Evaluation}
\label{sec:extrinsic}

Ultimately, our work aims to be more than just a theoretical exercise, but also to enable greater control over how sensitive NLP systems are to denotation vs.\ connotation in downstream tasks. To this end, we construct an ad hoc information retrieval task. We compare a system built on top of $V\sub{pretrained}$ to systems built on top of $V\sub{deno}$ and $V\sub{conno}$ in terms of both the quality of the ranking and the ideological diversity represented among the top results.

\subsection{Setup}

We focus only on \textsc{PN Proxy}  for this evaluation since it best matches the setting where we would expect to apply these techniques in practice: (1) We cannot always assume access to discrete denotation labels. (2) Language in the PN corpus is strongly influenced by ideology (as shown in Figure~\ref{fig:large-clusters}). 

To generate a realistic set of queries, we start with $12$ seed words from our vocabulary, chosen based on a list of the most important election issues for Democrat and Republican voters according to a recent Gallup Poll\footnote{https://news.gallup.com/poll/244367/top-issues-voters-healthcare-economy-immigration.aspx}. This results in the following list: ``economy, healthcare, immigration, women's rights, taxes, wealth, guns, climate change, foreign policy, supreme court, tariffs, special counsel". Then, for each seed word, we take 5 left-leaning seeds to be the 5 nearest neighbors according to $V\sub{pretrained}$, filtered to words which occur at least $100$ times and for which at least 70\% of occurrences appeared in left-leaning articles. We similarly chose 5 right-leaning seeds. We then submit each partisan seed to the Bing Autosuggest API and retrieve 10 suggestions each. We manually filter the list of queries to remove those that do not reflect the intended word sense (e.g., ``VA" leading to queries about Virginia rather than the Veterans Administration) and those which are not well matched to our document collection (e.g., queries seeking dictionary definitions, job openings, or specific websites such as Facebook). Our final list contains 410 queries, 216 left-leaning and 194 right-leaning. Table~\ref{tab:sample_queries} shows several examples, the full list is included in the supplementary material.

\begin{table}[ht!]
\begin{tabular}{|p{.9\linewidth}|}
\hline
\small
\textbf{Wealth:}
 \textcolor{brown}{globalist agenda}
$\circ$ \textcolor{brown}{globalist leaders}
$\circ$ \textcolor{lunar}{extreme poverty rates}
$\circ$ \textcolor{lunar}{romneys ties to burisma} \hspace{8mm}
\textbf{Women's Rights:}
 \textcolor{brown}{title ix impact}
$\circ$ \textcolor{brown}{safe spaces and snowflakes}
$\circ$ \textcolor{lunar}{anti-choice zealots}
$\circ$ \textcolor{lunar}{marriage equality court case}
\textbf{Immigration:}
 \textcolor{brown}{illegal immigrants at southern border}
$\circ$ \textcolor{brown}{illegals caught voting 2016}
$\circ$ \textcolor{lunar}{drug policy fbi}
$\circ$ \textcolor{lunar}{opioid crisis afghanistan}
\\
\hline
\end{tabular}
\caption{Example right- and left-leaning queries generated using the procedure described.}
\label{tab:sample_queries}
\end{table}

\subsection{Models}

We generate a ranked list of documents for each query in a two-step manner: (1) We pre-select the 5,000 most relevant documents according to a traditional BM25 model~\citep{robertson1995okapi} with default parameters. (2) This initial set of documents is then ranked using DRMM~\citep{guo2016deep}, a neural relevance matching model for ad-hoc retrieval. We train our retrieval model on the MS MARCO collection~\citep{bajaj2016ms} of 550,000 queries and 8.8 million documents from Bing. To highlight the effect of pretrained vs.\ decomposed word embeddings, we freeze our word embeddings during retrieval model training. While (1) is purely based on TF-IDF style statistics and remains static for all compared conditions, (2) is repeated for every proposed word embedding. This results in a ranked list of the top 100 most relevant documents for each query and word embedding.

\subsection{Results}
We compare the results of the DRMM retrieval model using different word embeddings in terms of quality and diversity of viewpoints reflected in the ranked results. To measure diversity, we report the overall distribution of political leanings among the top 100 documents and the rank-weighted $\alpha$-nDCG~\citep{clarke2008novelty} diversity score. For $\alpha$-nDCG, higher values indicate a more diverse list of results whose political leanings are evenly distributed across result list ranks. To measure ranking quality, we take a sample of $10$ queries and collect top $10$ results returned by each model variant, for a total of $300$ query/document pairs. We shuffle the list of  pairs to avoid biasing ourselves, and manually label each pair for whether or not the document is relevant to the query. We report Precision@$10$ estimated based on these $10$ queries.

\begin{figure}[ht!]
    \centering
        \includegraphics[width=7cm]{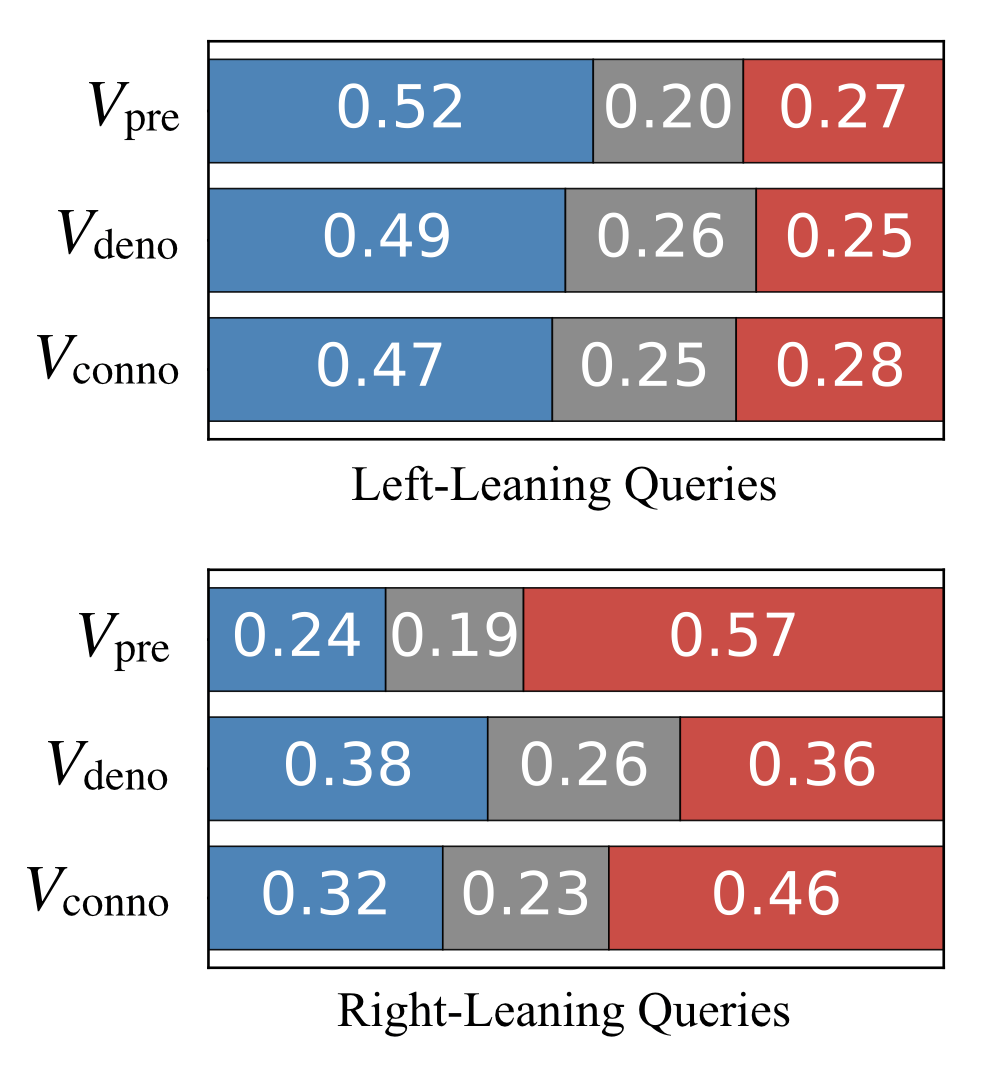}
    \caption{Distribution of partisanship of news source for top $100$ documents for right-leaning and left-leaning queries. Red = right-leaning news sources; blue = left-leaning; gray = nonpartisan or apolitical.}
    \label{fig:query-dist}
\end{figure}

Figure \ref{fig:query-dist} shows the overall party distributions. Table \ref{tab:extrinsic} reports the $\alpha$-nDCG and P@$10$ metrics. We can see that models which use $V\sub{deno}$ produce more diverse rankings  than do models that use $V\sub{pertained}$, with $V\sub{deno}$ producing an $\alpha$-nDCG@$100$ of $0.94$ vs.\ $0.92$ for pretrained. This trend is especially apparent in the rankings returned for right-leaning queries: Under the pretrained model, 57\% of the documents returned came from right-leaning news sources, whereas under the $V\sub{deno}$-based model, the results are nearly perfectly balanced between news sources. However, we do see a drop in precision when using $V\sub{deno}$. \hidden{(P@10 is 3.7 vs. 7.8 for pretrained)} This is not surprising given the limitations observed in \S\ref{sec:intrinsic}. If we had access to ground-truth denotation labels when training $V\sub{deno}$, we might expect to see these numbers improve. This is a promising direction for future work. 

\begin{table}[ht!]
\small
\centering
\begin{tabular}{lccccc}
\toprule
	&   \multicolumn{2}{c}{$\alpha$-nDCG}	& \multicolumn{2}{c}{Gini} & \\
& @$10$ & @$100$ &	L & R  &  P@$10$ \\
\cmidrule(lr){2-3}\cmidrule(lr){4-5}\cmidrule(lr){6-6}
$V\sub{pretrained}$      & $0.907$ & $0.915$ & $0.215$ & $0.207$ & $0.78$ \\
$V\sub{deno}$     & $0.922$ & $0.944$ & $0.160$ & $0.080$ & $0.37$ \\
$V\sub{conno}$     &$0.904$ & $0.914$  & $0.147$ & $0.153$ & $0.64$ \\
\bottomrule
\end{tabular}
\caption{Retrieval metrics. For $\alpha$-nDCG, higher means more diverse; for Gini, lower means more diverse. }
\label{tab:extrinsic}
\end{table}

\section{Related Work}

\paragraph{Embedding Augmentation.} At the lexical level, there is substantial literature that supplements pretrained representations with desired information \citep{faruqui-etal-2015-retrofitting,bamman-etal-2014-distributed} or improves their interpretability \citep{murphy-etal-2012-learning,arora2018linear,lauretig-2019-identification}. However, existing works tend to focus on evaluating the dictionary definitions of words, less so on grounding words to specific real world referents and, to our knowledge, no major attempt yet in interpreting and manipulating the denotation and connotation dimensions of meaning as suggested by the semantic theories discussed in \S\ref{sec:formulation}. While we do not claim to do full justice to conceptual role semantics either, this paper furnishes a first attempt at implementing a school of semantics introduced by philosophers of language and increasingly popular among cognitive scientists.

\paragraph{Style Transfer.} At the sentence level, adversarial setups similar to ours have been previously explored for differentiating style and content. For example, \citet{romanov-etal-2019-adversarial,yamshchikov-etal-2019-decomposing,john-etal-2019-disentangled} converted informal English to formal English and Yelp reviews from positive to negative sentiment. The motivation for such models is primarily natural language generation and the personalization thereof \citep{li-etal-2016-persona}. Additionally, our framing in terms of Frege's sense and reference adds clarity to the sometimes ill-defined problems explored in style transfer (e.g., treating sentiment as ``style''). For example, ``she is an undocumented immigrant'' and ``she is an illegal alien'' have the same truth conditions but different connotations, whereas  ``the cafe is great'' and ``the cafe is terrible'' have different truth conditions.

\paragraph{Modeling Political Language.} There is a wealth of work on computational approaches for modeling political language \citep{glavas-etal-2019-computational}. Within NLP, such efforts tend to focus more on describing how language differs between political subgroups, rather than recognizing similarities in denotation across ideological stances, which is the primary goal of our work. For example, \citet{preotiuc-pietro-etal-2017-beyond,han-etal-2019-fallacy} attempt to predict a person's political ideology from their social media posts, \citet{sim-etal-2013-measuring} detect ideological trends present in political speeches, \citet{fulgoni-etal-2016-empirical} predict political leaning of news articles, and \citet{pado-etal-2019-sides} focuses on modeling the network structure of policy debates within society. Also highly related is work analyzing linguistic framing in news \citep{greene-resnik-2009-words,choi-etal-2012-hedge,baumer-etal-2015-testing}. 

\paragraph{Echo Chambers and Search.} The dangers of ideological ``echo chambers'' have received significant attention across NLP, information retrieval, and social science research communities. \citet{dori2015navigating} discuss the challenges of deploying information retrieval systems in controversial domains, and \citet{puschmann2019beyond} looks specifically at the effects of search personalization on election-related information. Many approaches have been proposed to improve the diversity of search results, typically by identifying search facets \textit{a priori} and then training a model to optimize for diversity  \citep{tintarev2018same, tabrizi2019perspective, lunardi2019representing}. In terms of linguistic analyses, \citet{rashkin-etal-2017-truth} and \citet{potthast-etal-2018-stylometric} analyze stylistic patterns that distinguish fake news from real news. \citet{duseja-jhamtani-2019-sociolinguistic} study linguistic patterns that distinguish whether individuals are within social media echo chambers.
    
\section{Summary}
In this paper, we describe the problem of pretrained word embeddings conflating denotation and connotation. We address this issue by introducing an adversarial network that explicitly represents the two properties as two different vector spaces. We confirm that our decomposed spaces encode the desired structure of denotation or connotation by both quantitatively measuring their homogeneity and qualitatively evaluating their clusters and their representation of well-known political euphemisms. Lastly, we show that our decomposed spaces are capable of improving the diversity of document rankings in an information retrieval task.

\section*{Acknowledgment}

We are grateful to Jesse Shapiro, Stephen Bach, Yongming Han, Tucker Berckmann, Daniel Smits, Jessica Forde, Dylan Ebert, Aaron Traylor, Charles Lovering,  and Roma Patel for comments and discussions on the (many) early drafts of this paper. This research was supported by the Google Faculty Research Awards Program.

\bibliography{references}

\begin{thebibliography}{43}
\expandafter\ifx\csname natexlab\endcsname\relax\def\natexlab#1{#1}\fi

\bibitem[{Arora et~al.(2018)Arora, Li, Liang, Ma, and
  Risteski}]{arora2018linear}
Sanjeev Arora, Yuanzhi Li, Yingyu Liang, Tengyu Ma, and Andrej Risteski. 2018.
\newblock \href {http://arxiv.org/abs/1601.03764} {Linear algebraic structure
  of word senses, with applications to polysemy}.

\bibitem[{Bajaj et~al.(2016)Bajaj, Campos, Craswell, Deng, Gao, Liu, Majumder,
  McNamara, Mitra, Nguyen et~al.}]{bajaj2016ms}
Payal Bajaj, Daniel Campos, Nick Craswell, Li~Deng, Jianfeng Gao, Xiaodong Liu,
  Rangan Majumder, Andrew McNamara, Bhaskar Mitra, Tri Nguyen, et~al. 2016.
\newblock Ms marco: A human generated machine reading comprehension dataset.
\newblock \emph{arXiv preprint arXiv:1611.09268}.

\bibitem[{Bamman et~al.(2014)Bamman, Dyer, and
  Smith}]{bamman-etal-2014-distributed}
David Bamman, Chris Dyer, and Noah~A. Smith. 2014.
\newblock \href {https://doi.org/10.3115/v1/P14-2134} {Distributed
  representations of geographically situated language}.
\newblock In \emph{Proceedings of the 52nd Annual Meeting of the Association
  for Computational Linguistics (Volume 2: Short Papers)}, pages 828--834,
  Baltimore, Maryland. Association for Computational Linguistics.

\bibitem[{Baumer et~al.(2015)Baumer, Elovic, Qin, Polletta, and
  Gay}]{baumer-etal-2015-testing}
Eric Baumer, Elisha Elovic, Ying Qin, Francesca Polletta, and Geri Gay. 2015.
\newblock \href {https://doi.org/10.3115/v1/N15-1171} {Testing and comparing
  computational approaches for identifying the language of framing in political
  news}.
\newblock In \emph{Proceedings of the 2015 Conference of the North {A}merican
  Chapter of the Association for Computational Linguistics: Human Language
  Technologies}, pages 1472--1482, Denver, Colorado. Association for
  Computational Linguistics.

\bibitem[{Block(1986)}]{Block1986}
Ned Block. 1986.
\newblock Advertisement for a semantics for psychology.
\newblock \emph{Midwest studies in philosophy}, 10:615--678.

\bibitem[{Carey(2009)}]{Carey2009}
Susan Carey. 2009.
\newblock \href {https://books.google.com/books?id=h9J2DwAAQBAJ} {\emph{The
  Origin of Concepts}}.
\newblock Oxford series in cognitive development. Oxford University Press.

\bibitem[{Choi et~al.(2012)Choi, Tan, Lee, Danescu-Niculescu-Mizil, and
  Spindel}]{choi-etal-2012-hedge}
Eunsol Choi, Chenhao Tan, Lillian Lee, Cristian Danescu-Niculescu-Mizil, and
  Jennifer Spindel. 2012.
\newblock \href {https://www.aclweb.org/anthology/W12-3809} {Hedge detection as
  a lens on framing in the {GMO} debates: A position paper}.
\newblock In \emph{Proceedings of the Workshop on Extra-Propositional Aspects
  of Meaning in Computational Linguistics}, pages 70--79, Jeju, Republic of
  Korea. Association for Computational Linguistics.

\bibitem[{Clarke et~al.(2008)Clarke, Kolla, Cormack, Vechtomova, Ashkan,
  B{\"u}ttcher, and MacKinnon}]{clarke2008novelty}
Charles~LA Clarke, Maheedhar Kolla, Gordon~V Cormack, Olga Vechtomova, Azin
  Ashkan, Stefan B{\"u}ttcher, and Ian MacKinnon. 2008.
\newblock Novelty and diversity in information retrieval evaluation.
\newblock In \emph{Proceedings of the 31st annual international ACM SIGIR
  conference on Research and development in information retrieval}, pages
  659--666.

\bibitem[{Dori-Hacohen et~al.(2015)Dori-Hacohen, Yom-Tov, and
  Allan}]{dori2015navigating}
Shiri Dori-Hacohen, Elad Yom-Tov, and James Allan. 2015.
\newblock Navigating controversy as a complex search task.
\newblock In \emph{SCST@ ECIR}. Citeseer.

\bibitem[{Duseja and Jhamtani(2019)}]{duseja-jhamtani-2019-sociolinguistic}
Nikita Duseja and Harsh Jhamtani. 2019.
\newblock \href {https://doi.org/10.18653/v1/W19-2109} {A sociolinguistic study
  of online echo chambers on twitter}.
\newblock In \emph{Proceedings of the Third Workshop on Natural Language
  Processing and Computational Social Science}, pages 78--83, Minneapolis,
  Minnesota. Association for Computational Linguistics.

\bibitem[{Faruqui et~al.(2015)Faruqui, Dodge, Jauhar, Dyer, Hovy, and
  Smith}]{faruqui-etal-2015-retrofitting}
Manaal Faruqui, Jesse Dodge, Sujay~Kumar Jauhar, Chris Dyer, Eduard Hovy, and
  Noah~A. Smith. 2015.
\newblock \href {https://doi.org/10.3115/v1/N15-1184} {Retrofitting word
  vectors to semantic lexicons}.
\newblock In \emph{Proceedings of the 2015 Conference of the North {A}merican
  Chapter of the Association for Computational Linguistics: Human Language
  Technologies}, pages 1606--1615, Denver, Colorado. Association for
  Computational Linguistics.

\bibitem[{Frege(1892)}]{frege1892sinn}
Gottlob Frege. 1892.
\newblock {\"U}ber sinn und bedeutung.
\newblock \emph{Zeitschrift f{\"u}r Philosophie und philosophische Kritik},
  100:25--50.

\bibitem[{Fulgoni et~al.(2016)Fulgoni, Carpenter, Ungar, and
  Preo{\c{t}}iuc-Pietro}]{fulgoni-etal-2016-empirical}
Dean Fulgoni, Jordan Carpenter, Lyle Ungar, and Daniel Preo{\c{t}}iuc-Pietro.
  2016.
\newblock \href {https://www.aclweb.org/anthology/L16-1591} {An empirical
  exploration of moral foundations theory in partisan news sources}.
\newblock In \emph{Proceedings of the Tenth International Conference on
  Language Resources and Evaluation ({LREC}'16)}, pages 3730--3736,
  Portoro{\v{z}}, Slovenia. European Language Resources Association (ELRA).

\bibitem[{Gasparri and Marconi(2019)}]{sep-word-meaning}
Luca Gasparri and Diego Marconi. 2019.
\newblock {Word Meaning}.
\newblock In Edward~N. Zalta, editor, \emph{The {Stanford} Encyclopedia of
  Philosophy}, fall 2019 edition. Metaphysics Research Lab, Stanford
  University.

\bibitem[{Gentzkow et~al.(2019)Gentzkow, Shapiro, and
  Taddy}]{gentzkow2019measuring}
Matthew Gentzkow, Jesse~M Shapiro, and Matt Taddy. 2019.
\newblock Measuring group differences in high-dimensional choices: method and
  application to congressional speech.
\newblock \emph{Econometrica}, 87(4):1307--1340.

\bibitem[{Glava{\v{s}} et~al.(2019)Glava{\v{s}}, Nanni, and
  Ponzetto}]{glavas-etal-2019-computational}
Goran Glava{\v{s}}, Federico Nanni, and Simone~Paolo Ponzetto. 2019.
\newblock \href {https://doi.org/10.18653/v1/P19-4004} {Computational analysis
  of political texts: Bridging research efforts across communities}.
\newblock In \emph{Proceedings of the 57th Annual Meeting of the Association
  for Computational Linguistics: Tutorial Abstracts}, pages 18--23, Florence,
  Italy. Association for Computational Linguistics.

\bibitem[{Greenberg and Harman(2005)}]{greenberg2005conceptual}
Mark Greenberg and Gilbert Harman. 2005.
\newblock Conceptual role semantics.

\bibitem[{Greene and Resnik(2009)}]{greene-resnik-2009-words}
Stephan Greene and Philip Resnik. 2009.
\newblock \href {https://www.aclweb.org/anthology/N09-1057} {More than words:
  Syntactic packaging and implicit sentiment}.
\newblock In \emph{Proceedings of Human Language Technologies: The 2009 Annual
  Conference of the North {A}merican Chapter of the Association for
  Computational Linguistics}, pages 503--511, Boulder, Colorado. Association
  for Computational Linguistics.

\bibitem[{Guo et~al.(2016)Guo, Fan, Ai, and Croft}]{guo2016deep}
Jiafeng Guo, Yixing Fan, Qingyao Ai, and W~Bruce Croft. 2016.
\newblock A deep relevance matching model for ad-hoc retrieval.
\newblock In \emph{Proceedings of the 25th ACM International on Conference on
  Information and Knowledge Management}, pages 55--64.

\bibitem[{Han et~al.(2019)Han, Lee, Lee, and Cha}]{han-etal-2019-fallacy}
Jiyoung Han, Youngin Lee, Junbum Lee, and Meeyoung Cha. 2019.
\newblock \href {https://doi.org/10.18653/v1/D19-5548} {The fallacy of echo
  chambers: Analyzing the political slants of user-generated news comments in
  {K}orean media}.
\newblock In \emph{Proceedings of the 5th Workshop on Noisy User-generated Text
  (W-NUT 2019)}, pages 370--374, Hong Kong, China. Association for
  Computational Linguistics.

\bibitem[{John et~al.(2019)John, Mou, Bahuleyan, and
  Vechtomova}]{john-etal-2019-disentangled}
Vineet John, Lili Mou, Hareesh Bahuleyan, and Olga Vechtomova. 2019.
\newblock \href {https://doi.org/10.18653/v1/P19-1041} {Disentangled
  representation learning for non-parallel text style transfer}.
\newblock In \emph{Proceedings of the 57th Annual Meeting of the Association
  for Computational Linguistics}, pages 424--434, Florence, Italy. Association
  for Computational Linguistics.

\bibitem[{Kiesel et~al.(2019)Kiesel, Mestre, Shukla, Vincent, Adineh, Corney,
  Stein, and Potthast}]{kiesel-etal-2019-semeval}
Johannes Kiesel, Maria Mestre, Rishabh Shukla, Emmanuel Vincent, Payam Adineh,
  David Corney, Benno Stein, and Martin Potthast. 2019.
\newblock \href {https://doi.org/10.18653/v1/S19-2145} {{S}em{E}val-2019 task
  4: Hyperpartisan news detection}.
\newblock In \emph{Proceedings of the 13th International Workshop on Semantic
  Evaluation}, pages 829--839, Minneapolis, Minnesota, USA. Association for
  Computational Linguistics.

\bibitem[{Klein(2020)}]{klein2020we}
Ezra Klein. 2020.
\newblock \emph{Why We're Polarized}.
\newblock Profile Books.

\bibitem[{Kripke(1972)}]{kripke1972naming}
Saul~A Kripke. 1972.
\newblock Naming and necessity.
\newblock In \emph{Semantics of natural language}, pages 253--355. Springer.

\bibitem[{Lauretig(2019)}]{lauretig-2019-identification}
Adam Lauretig. 2019.
\newblock \href {https://doi.org/10.18653/v1/W19-2102} {Identification,
  interpretability, and {B}ayesian word embeddings}.
\newblock In \emph{Proceedings of the Third Workshop on Natural Language
  Processing and Computational Social Science}, pages 7--17, Minneapolis,
  Minnesota. Association for Computational Linguistics.

\bibitem[{Lee(2009)}]{lee2009beyond}
Frances~E Lee. 2009.
\newblock \emph{Beyond ideology: Politics, principles, and partisanship in the
  US Senate}.
\newblock University of Chicago Press.

\bibitem[{Li et~al.(2016)Li, Galley, Brockett, Spithourakis, Gao, and
  Dolan}]{li-etal-2016-persona}
Jiwei Li, Michel Galley, Chris Brockett, Georgios Spithourakis, Jianfeng Gao,
  and Bill Dolan. 2016.
\newblock \href {https://doi.org/10.18653/v1/P16-1094} {A persona-based neural
  conversation model}.
\newblock In \emph{Proceedings of the 54th Annual Meeting of the Association
  for Computational Linguistics (Volume 1: Long Papers)}, pages 994--1003,
  Berlin, Germany. Association for Computational Linguistics.

\bibitem[{Lunardi(2019)}]{lunardi2019representing}
Gabriel~Machado Lunardi. 2019.
\newblock Representing the filter bubble: Towards a model to diversification in
  news.
\newblock In \emph{International Conference on Conceptual Modeling}, pages
  239--246. Springer.

\bibitem[{Mikolov et~al.(2013)Mikolov, Sutskever, Chen, Corrado, and
  Dean}]{Mikolov2013}
Tomas Mikolov, Ilya Sutskever, Kai Chen, Greg~S Corrado, and Jeff Dean. 2013.
\newblock \href
  {http://papers.nips.cc/paper/5021-distributed-representations-of-words-and-phrases-and-their-compositionality.pdf}
  {Distributed representations of words and phrases and their
  compositionality}.
\newblock In \emph{Advances in Neural Information Processing Systems 26}, pages
  3111--3119. Curran Associates, Inc.

\bibitem[{Murphy et~al.(2012)Murphy, Talukdar, and
  Mitchell}]{murphy-etal-2012-learning}
Brian Murphy, Partha Talukdar, and Tom Mitchell. 2012.
\newblock \href {https://www.aclweb.org/anthology/C12-1118} {Learning effective
  and interpretable semantic models using non-negative sparse embedding}.
\newblock In \emph{Proceedings of {COLING} 2012}, pages 1933--1950, Mumbai,
  India. The COLING 2012 Organizing Committee.

\bibitem[{Pad{\'o} et~al.(2019)Pad{\'o}, Blessing, Blokker, Dayanik, Haunss,
  and Kuhn}]{pado-etal-2019-sides}
Sebastian Pad{\'o}, Andre Blessing, Nico Blokker, Erenay Dayanik, Sebastian
  Haunss, and Jonas Kuhn. 2019.
\newblock \href {https://doi.org/10.18653/v1/P19-1273} {Who sides with whom?
  towards computational construction of discourse networks for political
  debates}.
\newblock In \emph{Proceedings of the 57th Annual Meeting of the Association
  for Computational Linguistics}, pages 2841--2847, Florence, Italy.
  Association for Computational Linguistics.

\bibitem[{Potthast et~al.(2018)Potthast, Kiesel, Reinartz, Bevendorff, and
  Stein}]{potthast-etal-2018-stylometric}
Martin Potthast, Johannes Kiesel, Kevin Reinartz, Janek Bevendorff, and Benno
  Stein. 2018.
\newblock \href {https://doi.org/10.18653/v1/P18-1022} {A stylometric inquiry
  into hyperpartisan and fake news}.
\newblock In \emph{Proceedings of the 56th Annual Meeting of the Association
  for Computational Linguistics (Volume 1: Long Papers)}, pages 231--240,
  Melbourne, Australia. Association for Computational Linguistics.

\bibitem[{Preo{\c{t}}iuc-Pietro et~al.(2017)Preo{\c{t}}iuc-Pietro, Liu,
  Hopkins, and Ungar}]{preotiuc-pietro-etal-2017-beyond}
Daniel Preo{\c{t}}iuc-Pietro, Ye~Liu, Daniel Hopkins, and Lyle Ungar. 2017.
\newblock \href {https://doi.org/10.18653/v1/P17-1068} {Beyond binary labels:
  Political ideology prediction of twitter users}.
\newblock In \emph{Proceedings of the 55th Annual Meeting of the Association
  for Computational Linguistics (Volume 1: Long Papers)}, pages 729--740,
  Vancouver, Canada. Association for Computational Linguistics.

\bibitem[{Puschmann(2019)}]{puschmann2019beyond}
Cornelius Puschmann. 2019.
\newblock Beyond the bubble: Assessing the diversity of political search
  results.
\newblock \emph{Digital Journalism}, 7(6):824--843.

\bibitem[{Putnam(1975)}]{putnam1975meaning}
Hilary Putnam. 1975.
\newblock The meaning of ‘meaning’.
\newblock \emph{Philosophical papers}, 2.

\bibitem[{Qi et~al.(2020)Qi, Zhang, Zhang, Bolton, and
  Manning}]{qi-2020-stanza}
Peng Qi, Yuhao Zhang, Yuhui Zhang, Jason Bolton, and Christopher~D. Manning.
  2020.
\newblock \href {https://nlp.stanford.edu/pubs/qi2020stanza.pdf} {Stanza: A
  {Python} natural language processing toolkit for many human languages}.
\newblock In \emph{Proceedings of the 58th Annual Meeting of the Association
  for Computational Linguistics: System Demonstrations}.

\bibitem[{Rashkin et~al.(2017)Rashkin, Choi, Jang, Volkova, and
  Choi}]{rashkin-etal-2017-truth}
Hannah Rashkin, Eunsol Choi, Jin~Yea Jang, Svitlana Volkova, and Yejin Choi.
  2017.
\newblock \href {https://doi.org/10.18653/v1/D17-1317} {Truth of varying
  shades: Analyzing language in fake news and political fact-checking}.
\newblock In \emph{Proceedings of the 2017 Conference on Empirical Methods in
  Natural Language Processing}, pages 2931--2937, Copenhagen, Denmark.
  Association for Computational Linguistics.

\bibitem[{Robertson et~al.(1995)Robertson, Walker, Jones, Hancock-Beaulieu,
  Gatford et~al.}]{robertson1995okapi}
Stephen~E Robertson, Steve Walker, Susan Jones, Micheline~M Hancock-Beaulieu,
  Mike Gatford, et~al. 1995.
\newblock Okapi at trec-3.
\newblock \emph{Nist Special Publication Sp}, 109:109.

\bibitem[{Romanov et~al.(2019)Romanov, Rumshisky, Rogers, and
  Donahue}]{romanov-etal-2019-adversarial}
Alexey Romanov, Anna Rumshisky, Anna Rogers, and David Donahue. 2019.
\newblock \href {https://doi.org/10.18653/v1/N19-1088} {Adversarial
  decomposition of text representation}.
\newblock In \emph{Proceedings of the 2019 Conference of the North {A}merican
  Chapter of the Association for Computational Linguistics: Human Language
  Technologies, Volume 1 (Long and Short Papers)}, pages 815--825, Minneapolis,
  Minnesota. Association for Computational Linguistics.

\bibitem[{Sim et~al.(2013)Sim, Acree, Gross, and
  Smith}]{sim-etal-2013-measuring}
Yanchuan Sim, Brice D.~L. Acree, Justin~H. Gross, and Noah~A. Smith. 2013.
\newblock \href {https://www.aclweb.org/anthology/D13-1010} {Measuring
  ideological proportions in political speeches}.
\newblock In \emph{Proceedings of the 2013 Conference on Empirical Methods in
  Natural Language Processing}, pages 91--101, Seattle, Washington, USA.
  Association for Computational Linguistics.

\bibitem[{Tabrizi and Shakery(2019)}]{tabrizi2019perspective}
Shayan~A Tabrizi and Azadeh Shakery. 2019.
\newblock Perspective-based search: a new paradigm for bursting the information
  bubble.
\newblock \emph{FACETS}, 4(1):350--388.

\bibitem[{Tintarev et~al.(2018)Tintarev, Sullivan, Guldin, Qiu, and
  Odjik}]{tintarev2018same}
Nava Tintarev, Emily Sullivan, Dror Guldin, Sihang Qiu, and Daan Odjik. 2018.
\newblock Same, same, but different: algorithmic diversification of viewpoints
  in news.
\newblock In \emph{Adjunct Publication of the 26th Conference on User Modeling,
  Adaptation and Personalization}, pages 7--13.

\bibitem[{Yamshchikov et~al.(2019)Yamshchikov, Shibaev, Nagaev, Jost, and
  Tikhonov}]{yamshchikov-etal-2019-decomposing}
Ivan~P. Yamshchikov, Viacheslav Shibaev, Aleksander Nagaev, J{\"u}rgen Jost,
  and Alexey Tikhonov. 2019.
\newblock \href {https://doi.org/10.18653/v1/D19-5613} {Decomposing textual
  information for style transfer}.
\newblock In \emph{Proceedings of the 3rd Workshop on Neural Generation and
  Translation}, pages 128--137, Hong Kong. Association for Computational
  Linguistics.

\end{thebibliography}
\bibliographystyle{acl_natbib}

\newpage
\appendix

\section{Hyperparameters}
\begin{itemize}
    \item All classifier probes are 4-layer MLPs with hidden size 300, ReLU as nonlinearity, and dropout with $p$ = 0.33. 
    \item Decomposers $D$ and $C$ are embedding matrices of shape (vocab\_size, 300). Recomposer $R$ concatenates denotation and connotation as a 600-dimensional vector and then feed it into a linear layer of size (600, 300).
    \item The skip-gram loss follows the parameters recommended by \citet{Mikolov2013}. Context window radius = 5. Negative samples per true context word = 10. We also subsample frequent words in exactly the same way as the original paper (equation 5) did with their threshold of $10^{-5}$.
     \item We use Adam as our optimizer throughout. Learning rate = $1 \cross 10^{-3}$ for homogeneity and $1\cross 10^{-5}$ for Luntz-esque models. Other parameters left as PyTorch default.
    \item We train 30 epochs for large corpora (\textsc{CR Proxy}  and \textsc{PN Proxy} ). 150 epochs for smaller corpora (\textsc{CR Topic} and \textsc{CR Bill}).
    \item With batch size = 1024, the smaller corpora take about half an hour to train on an RTX 2080 Ti or comparable GPUs. With batch size = 8192, The larger corpora take about 50 hours to train.
    \item PyTorch version = 1.6. CUDA version = 10.2.
\end{itemize}

\begin{table}[ht!]
\small
\begin{tabular}{|p{.9\linewidth}|}
\hline
{\bf Luntz-esque:} estate tax, death tax, capitalism, free market, undocumented, illegal aliens, foreign trade, international trade, public option, governmentrun, political speech, campaign spending, cut taxes, trickledown, 
{\bf Random (CR):} cerro, brownfields, redtape, soon as possible, implicit, sup, habits, granted, personality, luis, internationally, itemize, fidel castro, centralize, restraint, pleadings, amendment before us, child custody protection, cheney, illegal aliens, 
{\bf Random  (PN):} reigniting, hurst, see happen, wandering, wp, conveying, obama obama, global politics, really serious, faggot, permanent normal, syrian observatory, native american, strength among, orbiting, protege, exclaimed, tunis, snopes staff, administration also, 
{\bf High Partisan (CR):} the usa patriot act, mining, patterns, public safety, gorge, spills, wall street, joliet, bridges, tax code, registrants, freedom of speech, compensatory time, college education, shelter, hunger, oil companies, scourge, somalia, traders, 
{\bf High Partisan (PN):} mrs. romney, pesticides, zionists, u.s. support, pacific northwest, economics defense, light bulbs, east asian, burton, smog, abdel fattah, banksters, work requirements, greenhouse gases, duggars, nigeria security, bolling, geopolitics, teng, newsom said \\
\hline
\end{tabular}
\caption{Sample words from each of our test sets as described in \S\ref{sec:test-sets}.}  
\label{tab:eval_sets}
\end{table}

\section{Preprocessing Procedures for Congressional Record}
\label{apd:preproc}
We use Stanford Stanza \cite{qi-2020-stanza} for tokenization, part-of-speech tag, dependency parsing, and named entity recognition. We replace multi-word phrases with an atomic token. We source our phrases of interests from the following three pipelines:
\begin{enumerate}
    \item Named entity recognizer.
    \item Frequency-based collocation. (We experimented with PMI-based collocation, which yielded results that were more prone to artifacts and arbitrary threshold setting.)
    \item Bigram and trigram constituents of parse trees that are (a) POS-tagged as noun phrase or verb phrase; (b) contain no stop words as in \code{nltk.corpus.stop\_words}; (c) contain no parliamentary procedural words as in \{“yield", “motion", “order", “ordered", “quorum", “roll", “unanimous", “mr.", “madam", “speaker", “chairman", “president", “senator",
        “gentleman", “colleague", “colleagues"\}
\end{enumerate}
From these sources, we filter vocabulary with minimum frequency = 15 for small corpora, 30 for large corpora. We then replace each phrase in the corpus by their respective tokens joined by an underscore. When words can be replaced by multiple phrases, longer phrases take priority, and then more frequent phrases take second priority.

Finally, we discard sentences with less than 5 words. We truncate sentences more than 20 words.

\section{Preprocessing Procedures for Partisan News}
\citet{kiesel-etal-2019-semeval} includes 600k articles for train and 150k articles for validation, each labeled with a 5-way partisanship by their publisher. We only train on their validation set because it is comparable in size with Congressional Record and it requires less data cleaning. We discard duplicate sentences, and the rest of the processing pipeline is the same as the Congressional Record.

As mentioned in the main paper, we find the corpus-given “left" vs. “left-center" and “right" and “right-center" labels are prone to artifacts of particular publishers. For example, many foreign policy related phrases dominate the “right-center" category simply because the publisher \italic{Foreign Affairs} is labeled as “right-center", but this distinction is unsupported in ground truth. Therefore, we collapse “left-center" and “left" as one class, and we collapse “right-center" and “right" as one class.

\section{Grounding Bill Titles and Topics}
\label{apd:bill-mentions}

We first filter out bills that are mentioned less than 3 times in its corresponding two-year congressional session. The vast majority of bills are only mentioned one time (when they were introduced) or twice (often a bipartisanship poster-child co-sponsor repeats the spiel.)

After manual inspection, we define three speeches after the bill mentioned speech as context speeches and thus assigned the same denotation label (bill or topic) as the bill mentioned speech. Statistics of bill mentioned for each congressional session is summarized in Table~\ref{tab:search-stat}. Subsequent tables show examples of bill topics, their frequency, and example bill mentioned speeches.

\begin{table*}[]
\centering
\begin{tabular}{@{}rrrrrr@{}}
\toprule
Session & Bills Scraped & \begin{tabular}[c]{@{}r@{}}Bill Title \\ RegEx Matches\end{tabular} & \begin{tabular}[c]{@{}r@{}}Bills with \\ $\geq$ 3 mentions\end{tabular} & \begin{tabular}[c]{@{}r@{}}Speeches with those \\ Bills Mentioned\end{tabular} & \begin{tabular}[c]{@{}r@{}}Num. Sentences \\ \end{tabular} \\ \midrule
97      & 1471          & 539                                                                 & 43                                                                 & 464                                                                            & 20372                                                                             \\
98      & 1633          & 688                                                                 & 51                                                                 & 665                                                                            & 33242                                                                              \\
99      & 1895          & 360                                                                 & 45                                                                 & 273                                                                            & 16128                                                                             \\
100     & 2092          & 440                                                                 & 47                                                                 & 358                                                                            & 18376                                                                             \\
101     & 2633          & 805                                                                 & 82                                                                 & 684                                                                            & 35903                                                                             \\
102     & 2778          & 626                                                                 & 58                                                                 & 503                                                                            & 26944                                                                             \\
103     & 2261          & 443                                                                 & 42                                                                 & 325                                                                            & 16500                                                                             \\
104     & 2120          & 548                                                                 & 46                                                                 & 440                                                                            & 21664                                                                             \\
105     & 2587          & 1174                                                                & 97                                                                 & 931                                                                            & 51878                                                                             \\
106     & 3421          & 1317                                                                & 115                                                                & 1033                                                                           & 64605                                                                             \\
107     & 3225          & 1007                                                                & 92                                                                 & 752                                                                            & 44901                                                                             \\
108     & 3039          & 688                                                                 & 75                                                                 & 436                                                                            & 26783                                                                             \\
109     & 3363          & 817                                                                 & 62                                                                 & 616                                                                            & 31838                                                                             \\
110     & 3928          & 1052                                                                & 102                                                                & 865                                                                            & 41601                                                                             \\
111     & 3714          & 868                                                                 & 73                                                                 & 740                                                                            & 36026                                                                             \\                                                              &                                                                                   \\ \bottomrule
\end{tabular}
\caption{Corpus with regular expression search for bill titles.}
\label{tab:search-stat}
\end{table*}

\begin{table*}
\centering
\begin{tabular}{ll} 
\toprule
Example Topic                                                                                   & Example Bill Short Titles                                        \\ \hline
\multirow{3}{*}{Health}                                                                         & National Diabetes Act                                \\
                                                                                                & Medical Devices Safety Act                           \\
                                                                                                & Emergency Medical Services Systems Act               \\ \hline
\multirow{3}{*}{Education}                                                                      & Women's Educational Equity Act                       \\
                                                                                                & Elementary and Secondary Drug Abuse Eradication Act  \\
                                                                                                & Community Education Development Act                  \\ \hline
\multirow{3}{*}{\begin{tabular}[c]{@{}l@{}} Government \\Operations and\\Politics\end{tabular}} & Nonpartisan Commission on Campaign Reform Act        \\
                                                                                                & Government in the Sunshine Act                       \\
                                                                                                & Congressional Disclosure of Income Act               \\
\bottomrule
\end{tabular}
\caption{Example bill topics.}
\end{table*}

\begin{table*}
\centering
\begin{tabular}{ll} 
\toprule
Freq. per sentence & Topic                                        \\ 
\hline
45815                  & Health                                       \\
38339                  & Education                                    \\
33993                  & Government operations and politics           \\
33462                  & Labor and employment                         \\
28392                  & Taxation                                     \\
26435                  & Crime and law enforcement                    \\
24204                  & Finance and financial sector                 \\
22273                  & Commerce                                     \\
21451                  & Transportation and public works              \\
20865                  & International affairs                        \\
18560                  & Public lands and natural resources           \\
17369                  & Armed forces and national security           \\
16376                  & Economics and public finance                 \\
15660                  & Law                                          \\
14702                  & Environmental protection                     \\
14472                  & Foreign trade and international finance      \\
13353                  & Families                                     \\
11752                  & Energy                                       \\
11741                  & Agriculture and food                         \\
10512                  & Science, technology, communications          \\
7050                   & Civil rights and liberties, minority issues  \\
6599                   & Housing and community development            \\
6066                   & Social welfare                               \\
5019                   & Native Americans                             \\
3582                   & Water resources development                  \\
3566                   & Commemorations                               \\
3457                   & Emergency management                         \\
2160                   & Immigration                                  \\
2116                   & Congress                                     \\
1640                   & Animals                                      \\
1559                   & Sports and recreation                        \\
1303                   & Day care                                     \\
552                    & Arts, culture, religion                      \\
545                    & Awards, medals, prizes                       \\
473                    & Public works                                 \\
389                    & Federal aid to handicapped services          \\
344                    & Monuments and memorials                      \\
241                    & Administrative procedure                     \\
157                    & Arms control                                 \\
123                    & Mines and mineral resources                  \\
94                     & Fires                                        \\
\bottomrule
\end{tabular}
\caption{\textsc{CR Topic}}
\end{table*}

\begin{table*}
\centering
\begin{tabular}{p{6in}} 
\toprule
Example Speeches with Bill Mentions \\                                                                                                                                                                                                                                                                                                                                                                                                                                                                                                                                              
\midrule 
\bold{“Auto Stock for Every Taxpayer Act”} These companies did all of this when the main company decided that the subsidiary was not consistent with the core business. That is what we should do with General Motors--give taxpayers its shares and get General Motors back in the marketplace where it belongs. This idea is fast. it is simple. and it creates a market for the shares... I ask unanimous consent to have printed in the RECORD newspaper articles supporting the Auto Stock for Every Taxpayer Act. \\  
\bold{“Radioactive Import Deterrence Act”} Mr. Speaker. the Radioactive Import Deterrence Act is a bipartisan bill that would ban the importation of lowlevel radioactive waste unless the President provides a waiver. Lowlevel radioactive waste is generated by medical facilities. university research labs. and utility companies. This waste is generated all over the United States. but finding permanent disposal sites has proven difficult. Currently. 36 States and the District of Columbia have only one approved site to store all the waste generated by those industries. That site is located in Utah... \\
\bold{“Help Find the Missing Act"}    I yield myself such time as I may consume. Madam Speaker. the Help Find the Missing Act. or Billys Law. will help families of missing persons find their loved ones by strengthening Federal databases about missing persons and unidentified remains. Every year. tens of thousands of Americans go missing and are never found. In the subcommittee we heard moving testimony from Ms. Janice Smolinski. whose son. Billy. went missing in 2004. While she has not found her son. she has dedicated her life to improving the system for others. including highlighting the need to strengthen and expand access to our missing persons databases. I thank her for her dedication to this worthy cause... \\
\bold{“Emergency Aid to American Survivors of the Haiti Earthquake Act"} Madam Speaker. I yield myself such time as I may consume. I rise in support of this Senate bill. S. 2949. As Representative MCDERMOTT described. it will provide assistance to thousands of Americans returning from Haiti following the devastating January 12 earthquake there. Let me reiterate that we are helping American citizens with this legislation. The bill. entitled Emergency Aid to American Survivors of the Haiti Earthquake Act. will ensure that State and local governments and charitable agencies on the ground in Florida... \\
\bold{“Enhanced Oversight of State and Local Economic Recovery Act"} Mr. Speaker. I rise to thank my colleagues for favorable consideration of H.R. 2182. the Enhanced Oversight of State and Local Economic Recovery Act. I was pleased to cosponsor this legislation. which was introduced by the chairman of the Oversight and Government Reform Committee. At a hearing of that committee. we learned that dedicated oversight funding for State and local governments could improve oversight of money appropriated through the American Recovery and Reinvestment Act...  \\                                                                                                                                                                                                                                                                                                                                                                                                                                                                                                                
\bold{“Veterans Dog Training Therapy Act"} I yield myself such time as I may consume. Madam Speaker. I rise today in support of H.R. 3885. the Veterans Dog Training Therapy Act. I want to thank the ranking member of the Health Subcommittee. Congressman BROWN from South Carolina. for bringing us this legislation. Madam Speaker. we all recognize how damaging the invisible wounds of war can be. The need for effective treatments for posttraumatic stress disorder and for other conditions. such as depression and substance abuse. is apparent. I think. to all Americans. This act recognizes and meets this need by exploring an innovative and promising new form of treatment. using the training of service dogs as a therapeutic medium... \\
\bold{“Prevent Deceptive Census Look Alike Mailings Act"} Mr. Speaker. entering its 23rd decade. the U.S. Census is the longest running national census in the world. Our founders wrote it into the Constitution. because taking a fair count is an essential part of fair government. A comprehensive. accurate Census helps ensure that our common resources are distributed where they are most needed. so that our communities can get the roads. schools. and police protection that they need. Theres nothing partisan about that goal. Unfortunately. some groups have set out to deceive Americans by disguising their own private mailings as Census documents... \\
\bottomrule
\end{tabular}
\caption{Seven random samples of bill mentions from the 111th Congress. Speeches truncated to fit the table.}

\end{table*}

\end{document}